\documentclass[letterpaper]{article} 
\usepackage{aaai2026}  
\usepackage{times}  
\usepackage{helvet}  
\usepackage{courier}  
\usepackage[hyphens]{url}  
\usepackage{graphicx} 
\usepackage{natbib}  
\usepackage{caption} 
\usepackage{algorithm}
\usepackage{algorithmic}

\usepackage{amsmath,amsfonts}
\usepackage[caption=false,font=scriptsize,labelfont=sf,textfont=sf]{subfig}
\usepackage{booktabs}
\usepackage{multirow}

\usepackage{newfloat}
\usepackage{listings}
\DeclareCaptionStyle{ruled}{labelfont=normalfont,labelsep=colon,strut=off} 
\floatstyle{ruled}
\newfloat{listing}{tb}{lst}{}
\floatname{listing}{Listing}
\title{Decoupling Scene Perception and Ego Status: A Multi-Context Fusion Approach for Enhanced Generalization in End-to-End Autonomous Driving}
\author{
    Jiacheng Tang\textsuperscript{\rm 1},
    Mingyue Feng\textsuperscript{\rm 2},
    Jiachao Liu\textsuperscript{\rm 2},
    Yaonong Wang\textsuperscript{\rm 2},
    Jian Pu\textsuperscript{\rm 1}\thanks{Corresponding author.}
}
\affiliations{
    \textsuperscript{\rm 1}Fudan University\\
    \textsuperscript{\rm 2}Zhejiang Leapmotor Technology Co., Ltd.\\

    jiachengtang21@m.fudan.edu.cn, jianpu@fudan.edu.cn
}

\usepackage{bibentry}

\begin{document}

\maketitle

\begin{abstract}
Modular design of planning-oriented autonomous driving has markedly advanced end-to-end systems. However, existing architectures remain constrained by an over-reliance on ego status, hindering generalization and robust scene understanding. We identify the root cause as an inherent design within these architectures that allows ego status to be easily leveraged as a shortcut. Specifically, the premature fusion of ego status in the upstream BEV encoder allows an information flow from this strong prior to dominate the downstream planning module. To address this challenge, we propose AdaptiveAD, an architectural-level solution based on a multi-context fusion strategy. Its core is a dual-branch structure that explicitly decouples scene perception and ego status. One branch performs scene-driven reasoning based on multi-task learning, but with ego status deliberately omitted from the BEV encoder, while the other conducts ego-driven reasoning based solely on the planning task. A scene-aware fusion module then adaptively integrates the complementary decisions from the two branches to form the final planning trajectory. To ensure this decoupling does not compromise multi-task learning, we introduce a path attention mechanism for ego-BEV interaction and add two targeted auxiliary tasks: BEV unidirectional distillation and autoregressive online mapping. Extensive evaluations on the nuScenes dataset demonstrate that AdaptiveAD achieves state-of-the-art open-loop planning performance. Crucially, it significantly mitigates the over-reliance on ego status and exhibits impressive generalization capabilities across diverse scenarios.
\end{abstract}


\section{Introduction}
\label{sec:introduction}

The modular design of planning-oriented autonomous driving \cite{hu2023planning} offers a novel paradigm for end-to-end models, yet it inherits the persistent challenge of causal confusion \cite{muller2005off}, where models learn spurious correlations from benchmark data. A critical manifestation is the tendency for models to `drive by inertia' rather than `driving by sight', a shortcut that leads to catastrophic failures in novel or long-tail scenarios where historical momentum fails to predict the requisite future action. Existing mitigation efforts have largely followed two paths: data-centric strategies, like balanced sampling \cite{chen2024survey}, which mitigate dataset biases but do not alter the model's internal information flow; and regularization techniques, such as dropout and contrastive imitation learning \cite{cheng2024pluto}, which improve feature quality but risk exacerbating the difficulty of multi-task learning in complex end-to-end frameworks. These approaches focus on refining the inputs to the decision-making process, rather than restructuring the process itself.

\begin{figure}[t]
    \centering
    \subfloat[Pipeline of planning-oriented autonomous driving]{
        \includegraphics[width=.96\linewidth]{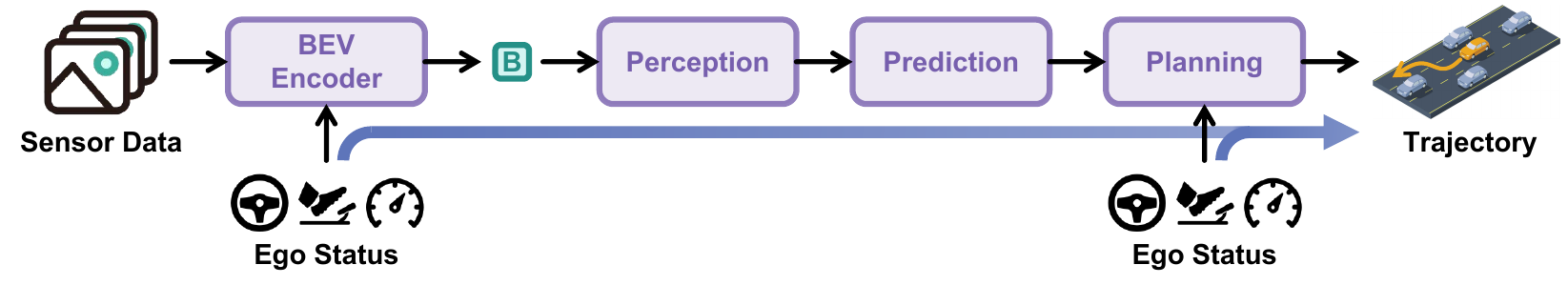}
        \label{fig:pipelinepoad}
    }

    \vspace{3mm}
    
    \subfloat[Pipeline of our AdaptiveAD]{
        \includegraphics[width=.96\linewidth]{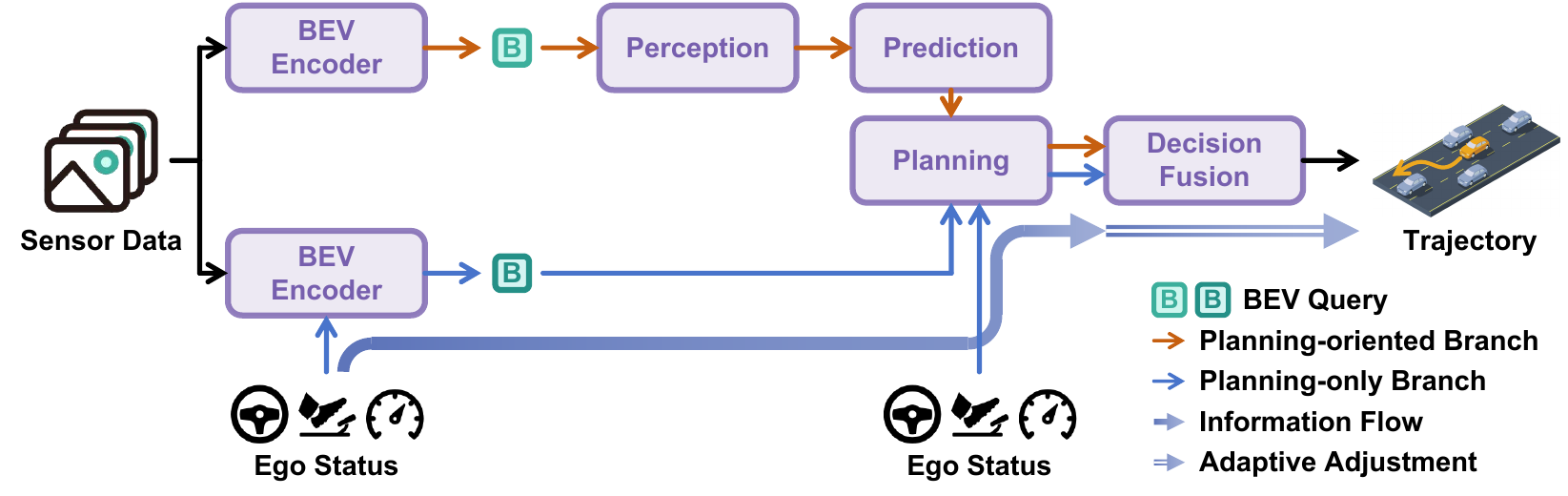}
        \label{fig:pipelineadaptivead}
    }
    \caption{
        Ego-status shortcut and our proposed architectural solution.
        (a) In conventional architectures, ego status is coupled with scene context, creating a shortcut that allows the planning module to rely on kinematic state.
        (b) Our AdaptiveAD framework uses a dual-branch design to explicitly decouple scene-driven reasoning from ego-status influence. A scene-aware fusion module then adaptively integrates these complementary decision contexts to generate the final trajectory.
    }
    \label{fig:pipeline}
\end{figure}

Recent investigations have revealed that this decision-making inertia stems from an over-reliance on the vehicle's own kinematic state, often termed ego status \cite{li2024ego}. For instance, when a high-speed vehicle must execute an emergency maneuver to avoid a sudden obstacle, a model `driving by inertia' is prone to generating a fatal trajectory. While dataset biases, such as the prevalence of straight-driving scenarios in benchmarks like nuScenes \cite{caesar2020nuscenes}, certainly exacerbate this issue, we argue the root cause is a critical but often overlooked inherent design that allows ego status to be easily leveraged as a shortcut. In many state-of-the-art architectures, ego status is fused with perceptual features early in the processing pipeline, as illustrated in Figure \ref{fig:pipelinepoad}. This unfiltered access allows the planning module to depend directly on ego status, bypassing complex scene understanding.

To address this, we propose AdaptiveAD, an architecture that explicitly decouples scene perception from ego status via a dual-branch, multi-context fusion strategy, as illustrated in Figure \ref{fig:pipelineadaptivead}. A scene-driven branch, with ego status excluded during BEV encoding, generates a planning decision derived solely from environmental cues. Concurrently, a lightweight, ego-driven branch generates a trajectory based primarily on the vehicle's kinematic state. These two complementary decision contexts—one encoding external scene constraints, the other the vehicle's own kinematic inertia—are then adaptively integrated by a scene-aware fusion module to produce a final trajectory that is both dynamically feasible and contextually appropriate.

Notably, architectural decoupling may introduce challenges for multi-task learning. To support our design, we further propose several key innovations. A novel path attention mechanism models the fine-grained interactions between potential ego paths and the dense BEV feature map, aiming to enhance both long-range dependency modeling and local detail capture. To further strengthen feature-behavior associations, we employ two auxiliary tasks. BEV unidirectional distillation is designed to mitigate potential motion blur in the perception features of the scene-driven branch that arise from the absence of ego status, while autoregressive online mapping aims to create a feedback loop from planning to mapping to improve multi-task learning efficacy. Evaluated on the nuScenes benchmark, AdaptiveAD not only achieves state-of-the-art open-loop planning performance but, more importantly, demonstrates significantly reduced reliance on the ego-status shortcut and superior generalization in complex scenarios.

Our primary contributions are outlined as follows:

\begin{itemize}
    \item We identify and address the architectural flaw that enables the ego-status shortcut. We propose a multi-context fusion strategy that explicitly decouples scene- and ego-driven reasoning, offering a robust solution to causal confusion in end-to-end models.
    \item We introduce three key innovations to support our decoupling strategy: a path attention mechanism to enhance ego-BEV interaction, BEV unidirectional distillation to preserve perceptual quality, and autoregressive online mapping to improve multi-task consistency.
    \item Our proposed framework, AdaptiveAD, sets a new state of the art for open-loop planning on the nuScenes dataset, demonstrating superior robustness and generalization in diverse and challenging driving scenarios.
\end{itemize}

\section{Related Work}
\label{sec:relatedwork}

\begin{figure*}[t]
    \centering
    \includegraphics[width=.96\linewidth]{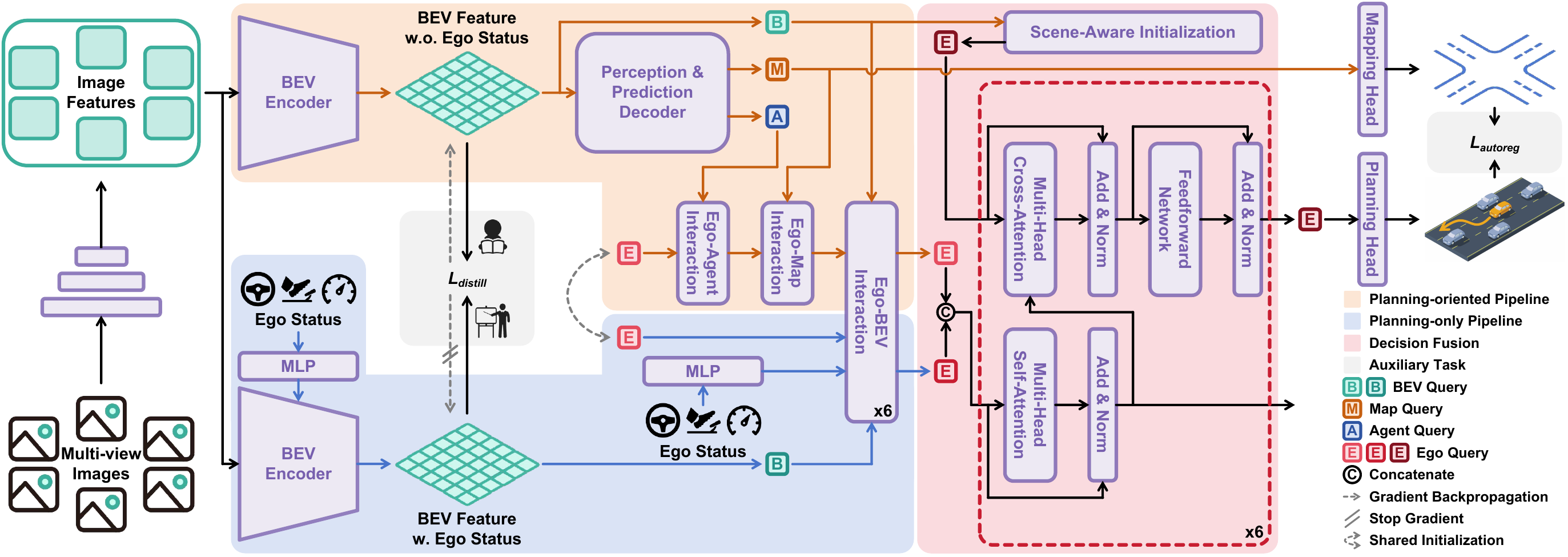}
    \caption{
        An overview of AdaptiveAD framework.
        Given a sequence of multi-view images, AdaptiveAD first extracts features using a shared backbone. The core of our framework is a dual-branch architecture that explicitly decouples information flow: one branch generates a scene-driven decision without ego-status influence, while a complementary branch produces an ego-driven decision. These distinct decision contexts are then adaptively integrated by a multi-context decision fusion module, which uses dense scene features as priors to generate the final trajectory. The integrity of this process is supported by two auxiliary tasks designed to enhance perceptual quality and enforce causal consistency.
    }
    \label{fig:adaptivead}
\end{figure*}

\subsection{End-to-End Autonomous Driving Architectures}
\label{subsec:e2eadarch}

The paradigm for autonomous driving has progressively shifted from modular pipelines, which are prone to error accumulation, towards integrated end-to-end systems. These systems aim to jointly optimize perception, prediction, and planning within a single differentiable framework. A significant milestone in this domain is the development of planning-oriented models like UniAD \cite{hu2023planning}, which leverage a unified query-based design to facilitate communication across different tasks. This approach has demonstrated that coordinating tasks with the ultimate goal of planning in mind can substantially improve performance over simple multi-task learning. However, while these architectures \cite{jiang2023vad} have advanced the field, their intricate data flow has inadvertently created new vulnerabilities, most notably the risk of learning spurious correlations from data biases \cite{li2024ego}.

\subsection{Causal Confusion and Ego-status Over-reliance}
\label{subsec:cc&esor}

Causal confusion, particularly shortcut learning from ego status \cite{li2024ego}, has prompted diverse mitigation strategies. Data-level methods, such as re-sampling \cite{chen2024survey} or augmentation \cite{katare2024analyzing}, alleviate dataset biases but do not fundamentally alter the model's internal information flow. Representation-level techniques, like dropout \cite{bansal2018chauffeurnet} and contrastive learning \cite{cheng2024pluto}, refine feature quality or restrict inputs to encourage perceptual reliance. While valuable, these approaches primarily address symptoms by refining inputs, rather than restructuring the decision-making process itself. In contrast, our work intervenes at the architectural level. We argue the root cause is the premature fusion of ego status within the BEV encoder, creating an information flow that allows the planner to bypass scene understanding \cite{li2024ego, yanlearning}. Instead of manipulating data or inputs, AdaptiveAD introduces a structural solution. By architecturally decoupling scene- and ego-driven reasoning before adaptive fusion, we directly suppress the ego-status shortcut, compelling the model to ground its decisions in environmental perception. This offers a more principled solution to causal confusion.



\subsection{Fusion Strategies and Feedback Mechanisms}
\label{subsec:dcf&fm}

Information fusion is central to robust autonomous systems. Multi-sensor fusion \cite{yang2022deepinteraction, cai2023bevfusion4d, yan2023cross, yan2024pointssc}, which combines data from different modalities like cameras and LiDAR, is a well-established technique for enhancing perception. However, its impact on planning is often indirect \cite{ye2023fusionad}. In contrast, our work explores the fusion of different decision contexts—a conceptually distinct approach aimed directly at improving the planning logic itself. Furthermore, enforcing causal consistency between a model's planned actions and its perception of the world is crucial for robust decision-making. Methodologies employing autoregressive mechanisms to create feedback loops have become influential for this purpose, ensuring that predicted future states align with intended actions \cite{li2024think2drive, gao2024vista}. Inspired by this principle, we introduce a targeted autoregressive online mapping task. This task establishes a feedback loop from planning to perception, enhancing the coherence and reliability of our multi-task learning architecture.


\section{Method}
\label{sec:method}

The core strategy of AdaptiveAD is to mitigate ego-status over-reliance through a principled, three-part architectural design. Our framework is built upon the explicit decoupling of information streams, the adaptive fusion of their outputs, and the support of targeted regularization to ensure robust multi-task learning. This design, illustrated in Figure \ref{fig:adaptivead}, systematically severs the shortcut pathway between ego status and planning. It begins by generating distinct scene-driven and ego-driven decisions in a dual-branch architecture, which are then intelligently integrated by a scenario-aware fusion module to produce a final, robust trajectory.

\subsection{Multi-context Decision Generation}
\label{subsec:mcdgeneration}

The cornerstone of AdaptiveAD is its dual-branch architecture, which processes perceptual inputs in two parallel streams, isolating the influence of ego-status information. This design allows the model to learn from both a pure, scene-centric representation and a conventional, ego-informed representation.

\subsubsection{Scene-driven Branch.}
\label{subsubsec:woesbranch}

This branch is engineered to generate planning decisions based exclusively on environmental perception, free from the direct influence of ego status. Following VAD \cite{jiang2023vad}, it comprises a BEV encoder, a vectorized scene decoder, and a decision generator that process multi-scale image features $F$. Our critical modification is removing the BEV query enhancement within the BEV encoder—a step that typically injects ego status and is a primary source of the shortcut. This removal yields a BEV feature, $B_{woes} \in \mathbb{R}^{C \times H_{bev} \times W_{bev}}$, representing the environment without ego-motion priors.

The subsequent vectorized scene decoder transforms this dense BEV into sparse agent queries $A \in \mathbb{R}^{N_{agent} \times C}$ and map queries $M \in \mathbb{R}^{N_{map} \times C}$. The decision generator then initializes a multimodal ego query $E_{woes} \in \mathbb{R}^{N_{mode} \times C}$ and uses it to interact sequentially with agent queries $A$, map queries $M$, and BEV feature $B_{woes}$ to generate a planning decision grounded solely in scene understanding.

\subsubsection{Planning-only Branch.}
\label{subsubsec:wesbranch}

In contrast, the ego-driven branch mirrors a more conventional architecture by incorporating ego status to generate a complementary decision. This streamlined branch retains the BEV query enhancement operation, producing a motion-compensated BEV feature map, $B_{wes} \in \mathbb{R}^{C \times H_{bev} \times W_{bev}}$, informed by ego-kinematics. It bypasses the explicit scene decoder; its ego query, $E_{wes} \in \mathbb{R}^{N_{mode} \times C}$, interacts directly with $B_{wes}$ to be updated into the final ego-driven decision. Reflecting a strong prior for motion extrapolation, the initial reference points for attention in this branch are predicted directly from the ego status.

\subsubsection{Path Attention.}
\label{subsubsec:pathattention}

\begin{figure}[t]
    \centering
    \includegraphics[width=.96\linewidth]{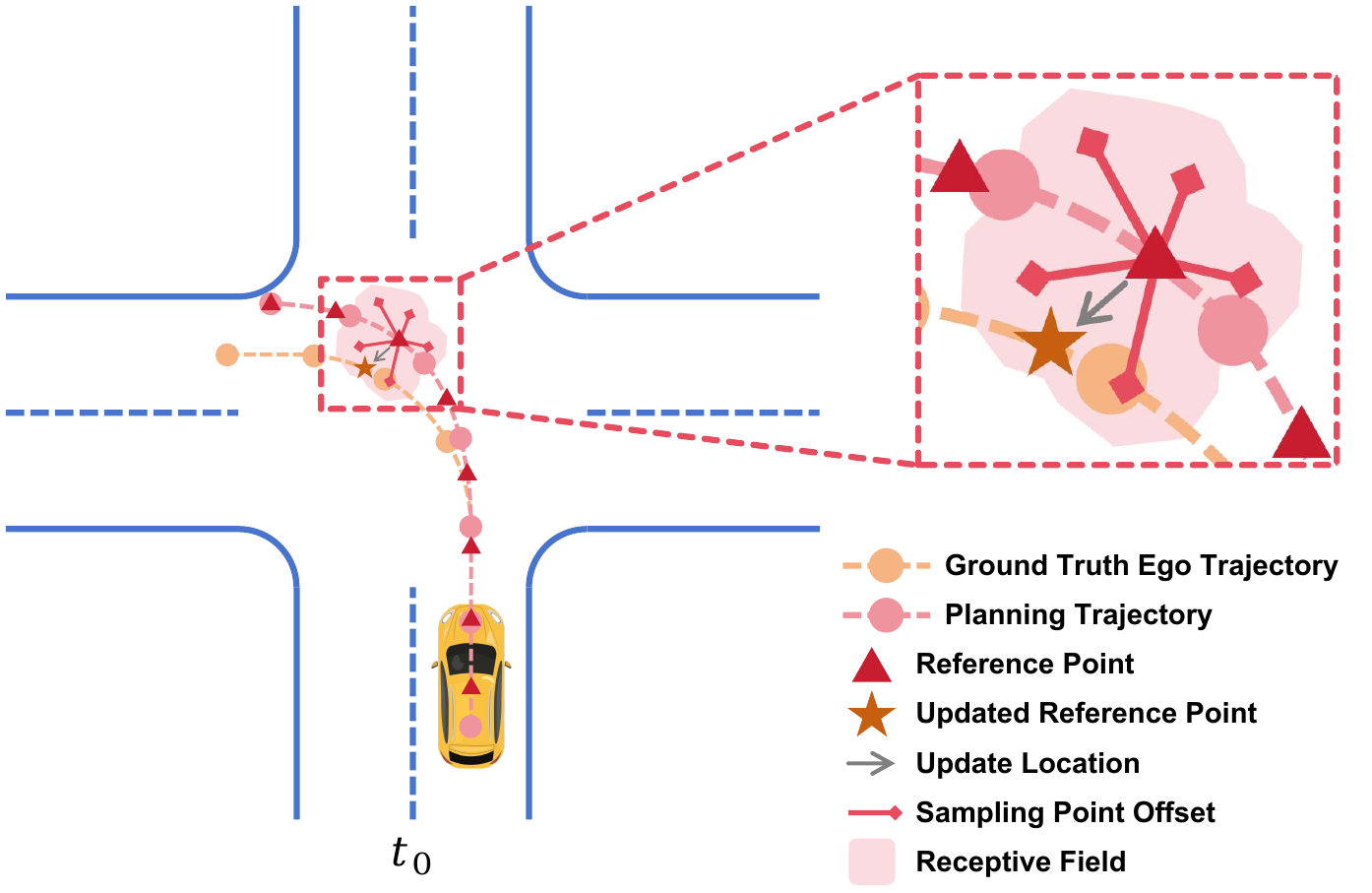}
    \caption{
        Diagram of path attention.
    }
    \label{fig:pathattention}
\end{figure}

Both branches leverage a novel path attention mechanism for the crucial interaction between the ego query and the BEV feature map. Standard deformable attention mechanisms learn to sample features from sparse, arbitrary locations, which is computationally efficient but lacks task-specific semantic grounding \cite{li2023lanesegnet}. Path attention refines this concept by introducing trajectory-guided semantic sampling, as depicted in Figure \ref{fig:pathattention}.

Specifically, for a given planning modality, we first decode a preliminary trajectory. We then sample $T$ reference points uniformly in time along this trajectory. Each reference point is assigned an independent attention head, which then learns to sample $K$ local features in its vicinity. This design constrains the model to gather evidence along a semantically meaningful axis—the hypothesized future path—mimicking how a human driver visually scans their intended route. For the $l$-th layer, given the $i$-th modality's ego query $E^{i} \in \mathbb{R}^{C}$ and its $T$ reference points $P^{i} \in \mathbb{R}^{T \times 2}$ on the BEV grid, path attention is formulated as:
\begin{equation}
    \text{PathAttn}(E^{i}, P^i, B) = \sum_{t=1}^{T} W_{t}[\sum_{k=1}^{K} a^{i,t,k} W'_{t} B_{samp}^{i,t,k}],
    \label{eq:pathattention}
\end{equation}
\begin{equation}
    B_{samp}^{i,t,k} = \text{BiLinear}(B, P^{i,t} + \Delta P^{i,t,k}).
    \label{eq:bilinear}
\end{equation}
where $B$ is the BEV feature map, $W'_{t} \in \mathbb{R}^{C_{T} \times C}$ and $W_{t} \in \mathbb{R}^{C \times C_{T}}$ are projection matrices for the $t$-th head, and $\text{BiLinear}(\cdot)$ performs bilinear sampling. The sampling offsets $\Delta P^{i,t,k}$ and attention weights $a^{i,t,k}$ are predicted from the ego query $E^{i}$. Crucially, the weights are normalized within each head ($\sum_{k=1}^{K} a^{i,t,k} = 1$), allowing the mechanism to exploit feature separation across heads to efficiently model both long-range context (via the spread of reference points) and local detail (via the learned offsets).

\subsection{Multi-context Decision Fusion}
\label{subsubsec:mcdfusion}

Our dual branches produce two distinct decision contexts: a scene-driven $E_{woes}$ for complex scenarios, and an ego-driven $E_{wes}$ providing a stable prior for inertial motion. As shown in Figure \ref{fig:adaptivead}, An adaptive fusion module then adaptively arbitrates between these complementary outputs.

First, to ground the final decision in the current environment, the fusion ego query $E_{fusion} \in \mathbb{R}^{N_{mode} \times C}$ is initialized with scene awareness. We extract a global scene feature from the scene-driven BEV map, which is shared across all planning modalities:
\begin{equation}
    E_{fusion}^{com} = \text{GAP}(B_{woes}),
    \label{eq:gap}
\end{equation}
where $\text{GAP}(\cdot)$ denotes global average pooling. This is combined with learnable modality-specific embeddings to form the initial $E_{fusion}$.

Next, a stack of transformer-based fusion layers merges the contexts. A key challenge is the potential feature space misalignment between $E_{woes}$ and $E_{wes}$. To address this, each fusion layer first performs a context alignment step. The two decision queries are concatenated and processed by a multi-head self-attention (MHSA) block:
\begin{equation}
    E_{multi} = \text{Concat}(\text{Proj}_{woes}(E_{woes}), \text{Proj}_{wes}(E_{wes})),
    \label{eq:concat}
\end{equation}
\begin{equation}
    E'_{multi} = \text{LayerNorm}(E_{multi} + \text{MHSA}(E_{multi})).
    \label{eq:resblock}
\end{equation}
This self-attention mechanism is critical, as it allows for rich inter-context interactions ($E_{woes} \rightarrow E_{wes}$, $E_{wes} \rightarrow E_{woes}$) and intra-context refinement ($E_{woes} \rightarrow E_{woes}$, $E_{wes} \rightarrow E_{wes}$). This step produces an aligned and enriched multi-context representation $E'_{multi} \in \mathbb{R}^{2N_{mode} \times C}$. The final decision query $E_{fusion}$ then attends to this representation via multi-head cross-attention to synthesize the final output, adaptively weighting the scene-driven and ego-driven information based on the demands of the current scenario.

\subsection{Auxiliary Tasks for Regularization}
\label{subsec:auxiliarytask}

Our dual-branch architecture is regularized by two targeted auxiliary tasks. These tasks preserve the quality of the decoupled representations and enforce consistency across the multi-task framework.

\subsubsection{BEV Unidirectional Distillation.}
\label{subsubsec:bevunidistill}

The scene-driven branch, by design, lacks ego-motion compensation, which can lead to motion blur in its BEV features ($B_{woes}$), particularly for dynamic agents. To counteract this, we introduce a BEV unidirectional distillation task. We treat the motion-compensated BEV from the ego-driven branch, $B_{wes}$, as a `teacher' and $B_{woes}$ as a `student'. This teacher-student paradigm is a proven method for knowledge transfer. The distillation loss is a composite objective:
\begin{equation}
    L_{distill} = \alpha L_{distill}^{DF} + \beta L_{distill}^{IK} + \gamma L_{distill}^{IC}.
    \label{eq:distillloss}
\end{equation}
Here, $L_{distill}^{DF}$ is a dense feature distillation loss with an agent-guided reweighting mechanism to focus on foreground objects. $L_{distill}^{IK}$ and $L_{distill}^{IC}$ are inter-keypoint and inter-channel distillation losses that encourage the student to learn feature correlations across agent keypoints and channels, narrowing the semantic gap between the two BEV representations. To ensure the teacher's stability, gradients are not propagated through $B_{wes}$ during this loss computation.

\subsubsection{Autoregressive Online Mapping.}
\label{subsubsec:aromapping}

End-to-end models that jointly perform mapping and planning can suffer from conflicting optimization goals, where the model might ambiguously learn to either move the planning trajectory away from a map element or move the map element away from the planning trajectory. To resolve this, we introduce an auxiliary task inspired by driving world models. World models learn to predict future states based on current states and actions, enforcing a causally consistent understanding of environmental dynamics.

\begin{figure}[t]
    \centering
    \includegraphics[width=.96\linewidth]{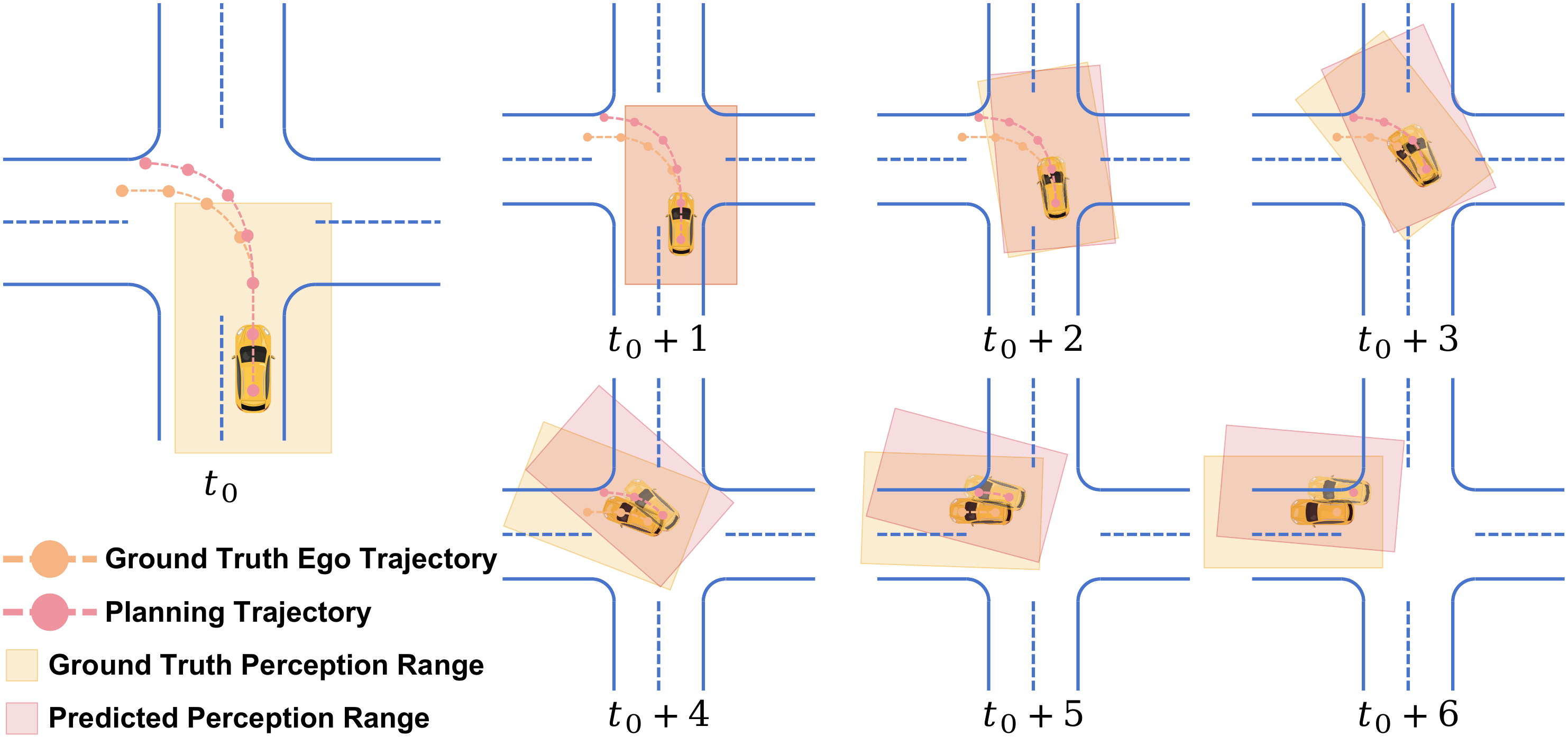}
    \caption{
        Diagram of autoregressive online mapping.
    }
    \label{fig:autoregression}
\end{figure}

Our task enforces that the perceived map should be consistent whether the ego vehicle follows the predicted trajectory or the ground-truth trajectory, as depicted in Figure \ref{fig:autoregression}. Let $\hat{P}_{E}, P_{E} \in \mathbb{R}^{T \times 2}$ be the predicted and ground-truth ego trajectories, and $\hat{P}_{M}, P_{M} \in \mathbb{R}^{N_{map} \times N_{point} \times 2}$ be the predicted and ground-truth map instances. For each future timestep $\tau$, we define an `autoregressive key region' as the intersection of the ego vehicle's perception bounding box if it were to follow $\hat{P}_{E}$ versus $P_{E}$. We then apply a masked L1 loss, $L_{autoreg}^{MAP}$, only on the map instances within this region:
\begin{equation}
    L_{autoreg}^{MAP} = \frac{1}{T} \sum_{\tau = 1}^{T} \frac{1}{\left\| \mathcal{M} \right\|_{1} + \epsilon} \left\| (\hat{P}_{M} - P_{M}) \odot \mathcal{M} \right\|_{1},
    \label{eq:mappingloss}
\end{equation}
where $\mathcal{M} = \text{Mask}(P_{M}, \hat{R}_{E}^{\tau} \cap R_{E}^{\tau})$ is the mask for key instances in the overlapping perception region, with $\epsilon$ being a smoothing term. This ensures the model learns a consistent mapping outcome conditioned on its planned actions. To handle cases with no overlap and provide a stable gradient, we add a Gaussian Wasserstein distance loss \cite{yang2021rethinking}, $L_{autoreg}^{GWD}$, on the perception bounding boxes themselves. The total loss is:
\begin{equation}
    L_{autoreg} = \delta L_{autoreg}^{MAP} + \lambda L_{autoreg}^{GWD}.
    \label{eq:autoregressiveloss}
\end{equation}
This autoregressive supervision establishes a feedback loop from planning to mapping, mitigating task conflict and promoting a more coherent world representation.

\section{Experiments}
\label{sec:experiments}

\subsection{Experimental Setup}

\subsubsection{Datasets and Evaluation Metrics.}

Our primary evaluation is conducted on the nuScenes \cite{caesar2020nuscenes} dataset, a standard benchmark for vision-centric autonomous driving. Following established protocols \cite{jiang2023vad}, we train on the train set and report results on the val set. We use the primary open-loop planning metrics: L2 Displacement Error (L2) to measure trajectory accuracy and Collision Rate (CR) to assess safety \cite{jiang2023vad}. To further probe model robustness, we conduct supplementary experiments on the NAVSIM \cite{dauner2024navsim} non-reactive benchmark and the Bench2Drive \cite{jia2024bench2drive} closed-loop simulator, using their official metrics: PDM Score (PDMS) \cite{dauner2024navsim}, Driving Score (DS), and Success Rate (SR) \cite{jia2024bench2drive}. Unless otherwise specified, all reported results are based on the nuScenes dataset. And all FPS measurements are conducted on one NVIDIA GeForce RTX 3090 GPU, with the exception of UniAD \cite{hu2023planning} and PPAD \cite{chen2024ppad}.

\subsubsection{Implementation Details.}

AdaptiveAD is implemented in PyTorch, leveraging the open-source MMDetection3D framework. Our configuration largely follows VAD, predicting a 3-second trajectory from 2 seconds of historical data within a 60m x 30m perception range. The model includes 6 ego-BEV interaction layers and 6 multi-context fusion layers. Loss weights for auxiliary tasks ($\alpha$, $\beta$, $\gamma$, $\delta$, $\lambda$) are set to (0.01, 0.1, 0.01, 0.01, 0.01). We train for 60 epochs on 32 NVIDIA A100 GPUs using the AdamW optimizer and a CosineAnnealing scheduler, with a batch size of 2 per GPU.

\begin{table}[H]
    \centering
    \small
    \setlength{\tabcolsep}{2pt} 
    \begin{tabular}{l|cccc|cccc|c}
        \specialrule{1.0pt}{0pt}{0pt}
        \multirow{2}{*}{\textbf{Method}} & \multicolumn{4}{c|}{\textbf{L2 (m) $\downarrow$}} & \multicolumn{4}{c|}{\textbf{Collision ($\%$) $\downarrow$}} & \multirow{2}{*}{\textbf{FPS}} \\
         & 1s & 2s & 3s & Avg. & 1s & 2s & 3s & Avg. &  \\
        \hline
        UniAD & 0.45 & 0.70 & 1.04 & 0.73 & 0.62 & 0.58 & 0.63 & 0.61 & 1.8 \\
        VAD $\S$ & 0.31 & 0.58 & 0.94 & 0.61 & 0.17 & 0.25 & 0.43 & 0.28 & 3.4 \\
        PPAD & 0.31 & 0.56 & 0.87 & 0.58 & 0.08 & 0.12 & 0.38 & 0.19 & 2.6 \\
        SparseDrive $\dag$ & 0.30 & 0.58 & 0.95 & 0.61 & 0.01 & 0.05 & 0.23 & 0.10 & 5.2 \\
        BridgeAD $\dag$ & 0.28 & 0.55 & 0.92 & 0.58 & \textbf{0.00} & \textbf{0.04} & 0.20 & \textbf{0.08} & 3.1 \\
        FusionAD & - & - & - & 1.03 & 0.25 & 0.13 & 0.25 & 0.21 & - \\
        \hline
        Ours & \textbf{0.23} & \textbf{0.43} & \textbf{0.74} & \textbf{0.47} & 0.05 & 0.12 & \textbf{0.18} & 0.12 & 3.0 \\
        \specialrule{1.0pt}{0pt}{0pt}
    \end{tabular}
    \caption{
        Open-loop planning performance.
        $\S$ denotes re-implement result. $\dag$ denotes that the auxiliary task of predicting ego status was used during training.
    }
    \label{tab:planningperformance}
\end{table}

\begin{table}[H]
    \centering
    \small
    \setlength{\tabcolsep}{3pt} 
    \begin{tabular}{l|c|cccc|cccc}
        \specialrule{1.0pt}{0pt}{0pt}
        \multirow{2}{*}{\textbf{Method}} & \multirow{2}{*}{\textbf{Nav.}} & \multicolumn{4}{c|}{\textbf{L2 (m) $\downarrow$}} & \multicolumn{4}{c}{\textbf{Collision ($\%$)} $\downarrow$} \\
         &  & 1s & 2s & 3s & Avg. & 1s & 2s & 3s & Avg. \\
        \hline
        \multirow{3}{*}{VAD} & ST & 0.32 & 0.59 & 0.95 & 0.62 & 0.20 & 0.30 & 0.48 & 0.33 \\
         & LR & 0.50 & 0.88 & 1.35 & 0.91 & \textbf{0.00} & \textbf{0.11} & 0.44 & 0.18 \\
         & LR $\ddag$ & 0.40 & 0.83 & 1.52 & 0.92 & 0.07 & 0.22 & 0.84 & 0.38 \\
        \hline
        \multirow{3}{*}{Ours} & ST & \textbf{0.23} & \textbf{0.43} & \textbf{0.74} & \textbf{0.47} & \textbf{0.06} & \textbf{0.11} & \textbf{0.17} & \textbf{0.11} \\
         & LR & \textbf{0.34} & \textbf{0.61} & \textbf{0.95} & \textbf{0.63} & \textbf{0.00} & 0.15 & \textbf{0.34} & \textbf{0.16} \\
         & LR $\ddag$ & \textbf{0.26} & \textbf{0.55} & \textbf{1.09} & \textbf{0.63} & \textbf{0.04} & \textbf{0.11} & \textbf{0.68} & \textbf{0.28} \\
        \specialrule{1.0pt}{0pt}{0pt}
    \end{tabular}
    \caption{
        Scene generalization ability.
        Navigation commands are labeled as `ST' (straight) and `LR' (left/right turn) \cite{li2024ego}. $\ddag$ split follows the Turning-nuScenes dataset's official protocol \cite{song2025don}.
    }
    \label{tab:scenegeneralization}
\end{table}

\begin{table}[H]
    \centering
    \small
    \setlength{\tabcolsep}{1.65pt} 
    \begin{tabular}{l|c|cccc|cccc}
        \specialrule{1.0pt}{0pt}{0pt}
        \multirow{2}{*}{\textbf{Method}} & \textbf{Velo.} & \multicolumn{4}{c|}{\textbf{L2 (m) $\downarrow$}} & \multicolumn{4}{c}{\textbf{Collision ($\%$) $\downarrow$}} \\
         & \textbf{Noise} & 1s & 2s & 3s & Avg. & 1s & 2s & 3s & Avg. \\
        \hline
        \multirow{5}{*}{VAD} & - & 0.31 & 0.58 & 0.94 & 0.61 & 0.17 & 0.25 & 0.43 & 0.28 \\
         & $\times 0.0$ & 3.53 & 5.61 & 7.48 & 5.54 & 0.53 & 2.15 & 3.85 & 2.18 \\
         & $\times 0.5$ & 1.90 & 3.07 & 4.17 & 3.05 & 0.23 & 0.52 & 1.15 & 0.63 \\
         & $\times 1.5$ & 1.91 & 3.21 & 4.53 & 3.22 & 0.22 & 0.96 & 1.98 & 1.05 \\
         & $100m/s$ & 8.94 & 14.92 & 20.93 & 14.93 & 7.47 & 6.20 & 5.11 & 6.26 \\
        \hline
        \multirow{5}{*}{Ours} & - & 0.23 & 0.43 & 0.74 & 0.47 & 0.05 & 0.12 & 0.18 & 0.12 \\
         & $\times 0.0$ & 2.62 & 4.18 & 5.43 & 4.08 & 0.20 & 1.07 & 2.13 & 1.13 \\
         & $\times 0.5$ & 1.52 & 2.45 & 3.26 & 2.41 & 0.09 & 0.16 & 0.49 & 0.25 \\
         & $\times 1.5$ & 1.66 & 2.76 & 3.81 & 2.74 & 0.16 & 0.60 & 1.53 & 0.76 \\
         & $100m/s$ & 2.89 & 5.04 & 7.24 & 5.06 & 3.25 & 5.42 & 5.92 & 4.86 \\
        \specialrule{1.0pt}{0pt}{0pt}
    \end{tabular}
    \caption{
        Reliance on ego status on nuScenes.
        We assign different levels of perturbation to the ego-velocity.
    }
    \label{tab:relianceonesonnus}
\end{table}

\begin{table}[H]
    \centering
    \small
    \setlength{\tabcolsep}{8.7pt} 
    \begin{tabular}{l|c|c|cc}
        \specialrule{1.0pt}{0pt}{0pt}
        \multirow{2}{*}{\textbf{Method}} & \textbf{Velo.} & \multicolumn{1}{c|}{\textbf{NAVSIM}} & \multicolumn{2}{c}{\textbf{Bench2Drive}} \\
         & \textbf{Noise} & PDMS $\uparrow$ & DS $\uparrow$ & SR($\%$) $\uparrow$ \\
        \hline
        \multirow{5}{*}{VAD} & - & 81.2 & 44.35 & 16.91 \\
         & $\times 0.0$ & 51.5 & 14.63 & 3.45 \\
         & $\times 0.5$ & 78.5 & 41.22 & 13.18 \\
         & $\times 1.5$ & 67.9 & 22.78 & 7.27 \\
         & $100m/s$ & 22.1 & 6.14 & 0.00 \\
        \hline
        \multirow{5}{*}{Ours} & - & 86.4 & 49.47 & 19.23 \\
         & $\times 0.0$ & 61.4 & 19.91 & 5.82 \\
         & $\times 0.5$ & 84.6 & 45.81 & 16.44 \\
         & $\times 1.5$ & 76.6 & 29.88 & 9.54 \\
         & $100m/s$ & 30.7 & 9.42 & 0.00 \\
        \specialrule{1.0pt}{0pt}{0pt}
    \end{tabular}
    \caption{
        Reliance on ego status on NAVSIM and Bench2Drive.
    }
    \label{tab:relianceonesonnavben}
\end{table}

\begin{table}[H]
    \centering
    \small
    \setlength{\tabcolsep}{1.57pt} 
    \begin{tabular}{c|cccc|cccc|cccc|c}
        \specialrule{1.0pt}{0pt}{0pt}
        \multirow{2}{*}{\textbf{ID}} & \multirow{2}{*}{\textbf{D.}} & \multirow{2}{*}{\textbf{B.}} & \multirow{2}{*}{\textbf{S.}} & \multirow{2}{*}{\textbf{A.}} & \multicolumn{4}{c|}{\textbf{L2 (m) $\downarrow$}} & \multicolumn{4}{c|}{\textbf{Collision ($\%$) $\downarrow$}} & \multirow{2}{*}{\textbf{FPS}} \\
         &  &  &  &  & 1s & 2s & 3s & Avg. & 1s & 2s & 3s & Avg. &  \\
        \hline
        1 & - & - & - & - & 0.28 & 0.54 & 0.90 & 0.57 & 0.10 & 0.20 & 0.35 & 0.22 & 3.4 \\
        2 & \checkmark & - & - & - & 0.30 & 0.58 & 0.98 & 0.62 & 0.10 & 0.14 & 0.20 & 0.15 & 3.0 \\
        3 & \checkmark & \checkmark & - & - & 0.28 & 0.53 & 0.92 & 0.58 & 0.03 & 0.05 & 0.17 & 0.08 & 3.0 \\
        4 & \checkmark & \checkmark & \checkmark & - & 0.25 & 0.49 & 0.82 & 0.52 & 0.06 & 0.13 & 0.18 & 0.12 & 3.0 \\
        5 & \checkmark & \checkmark & \checkmark & \checkmark & \textbf{0.23} & \textbf{0.43} & \textbf{0.74} & \textbf{0.47} & \textbf{0.05} & \textbf{0.12} & \textbf{0.18} & \textbf{0.12} & 3.0 \\
        \specialrule{1.0pt}{0pt}{0pt}
    \end{tabular}
    \caption{
        Ablation studies for innovative components.
        `D.' (dual-branch), `B.' (BEV unidirectional distillation), `S.' (scene-aware initialization), and `A.' (autoregressive online mapping) are incrementally added to the baseline.
    }
    \label{tab:ablationcomponents}
\end{table}

\begin{table}[H]
    \centering
    \small
    \setlength{\tabcolsep}{3.31pt} 
    \begin{tabular}{l|cccc|cccc|c}
        \specialrule{1.0pt}{0pt}{0pt}
        \multirow{2}{*}{\textbf{Mech.}} & \multicolumn{4}{c|}{\textbf{L2 (m) $\downarrow$}} & \multicolumn{4}{c|}{\textbf{Collision ($\%$) $\downarrow$}} & \multirow{2}{*}{\textbf{FPS}} \\
         & 1s & 2s & 3s & Avg. & 1s & 2s & 3s & Avg. &  \\
        \hline
        D.A. & 0.23 & 0.45 & 0.77 & 0.48 & 0.07 & 0.13 & 0.25 & 0.15 & 3.0 \\
        P.A. & \textbf{0.23} & \textbf{0.43} & \textbf{0.74} & \textbf{0.47} & \textbf{0.05} & \textbf{0.12} & \textbf{0.18} & \textbf{0.12} & 3.0 \\
        \specialrule{1.0pt}{0pt}{0pt}
    \end{tabular}
    \caption{
        Effectiveness of path attention.
    }
    \label{tab:pathattention}
\end{table}

\begin{table}[H]
    \centering
    \small
    \setlength{\tabcolsep}{3.46pt} 
    \begin{tabular}{l|cccc|cccc}
        \specialrule{1.0pt}{0pt}{0pt}
        \multirow{2}{*}{\textbf{Method}} & \multicolumn{4}{c|}{\textbf{L2 (m) $\downarrow$}} & \multicolumn{4}{c}{\textbf{Collision ($\%$) $\downarrow$}} \\
         & 1s & 2s & 3s & Avg. & 1s & 2s & 3s & Avg. \\
        \hline
        UniAD & 0.45 & 0.70 & 1.04 & 0.73 & 0.62 & 0.58 & 0.63 & 0.61 \\
        + P.A. & 0.43 & 0.65 & 0.96 & 0.68 & 0.57 & 0.55 & 0.54 & 0.55 \\
        \hline
        SparseDrive & 0.29 & 0.55 & 0.91 & 0.58 & 0.01 & 0.02 & 0.13 & 0.06 \\
        + A.O.M. & 0.27 & 0.49 & 0.82 & 0.53 & 0.01 & 0.03 & 0.14 & 0.06 \\
        \specialrule{1.0pt}{0pt}{0pt}
    \end{tabular}
    \caption{
        Effectiveness of plugins on open-loop planning.
    }
    \label{tab:pluginplanningperformance}
\end{table}

\subsection{Main Results}

\subsubsection{Open-loop Planning Performance.}

As shown in Table \ref{tab:planningperformance}, AdaptiveAD sets a new state of the art in planning accuracy on the nuScenes benchmark with the lowest average L2 error. It also achieves a highly competitive collision rate \cite{zhang2025bridging}, significantly outperforming our primary baseline, VAD, by reducing the average L2 error by 22$\%$ and the collision rate by 57$\%$. Notably, this performance is achieved with negligible impact on latency, maintaining an inference speed of 3.0 FPS.

\subsubsection{Scene Generalization Ability.}

The nuScenes dataset is heavily skewed towards simple, straight-driving scenarios ($\sim$75$\%$), which can mask model weaknesses. To dissect our performance, we evaluate generalization on complex turning maneuvers (LR) versus simple straight driving (ST), as shown in Table \ref{tab:scenegeneralization}. While both models perform well when driving straight, VAD's performance degrades significantly in turning scenarios. In contrast, AdaptiveAD demonstrates far more consistent performance, substantially outperforming VAD in both L2 error and CR during turns. This result highlights AdaptiveAD's superior ability to rely on scene understanding when ego-status priors become unreliable.

\subsubsection{Reliance on Ego Status.}

To directly test our core hypothesis, we inject varying levels of noise into the ego-velocity input during inference. As detailed in Table \ref{tab:relianceonesonnus}, VAD's performance collapses catastrophically under noisy ego status, with L2 error increasing by over 800$\%$ when velocity is zeroed out. AdaptiveAD, while still affected, exhibits markedly greater resilience, with significantly smaller performance degradation across all noise levels. This demonstrates that our multi-context fusion strategy successfully reduces the model's hazardous over-dependence on ego status. We confirm this finding in more realistic settings on NAVSIM and the closed-loop CARLA-based Bench2Drive (Table \ref{tab:relianceonesonnavben}), where AdaptiveAD consistently maintains higher and more stable driving scores under perturbation.

\subsection{Ablation Studies}

We conduct a series of ablation studies, summarized in Table \ref{tab:ablationcomponents}, to systematically validate the contribution of each component in AdaptiveAD.

\begin{figure*}[t]
    \centering
    \includegraphics[width=.75\linewidth]{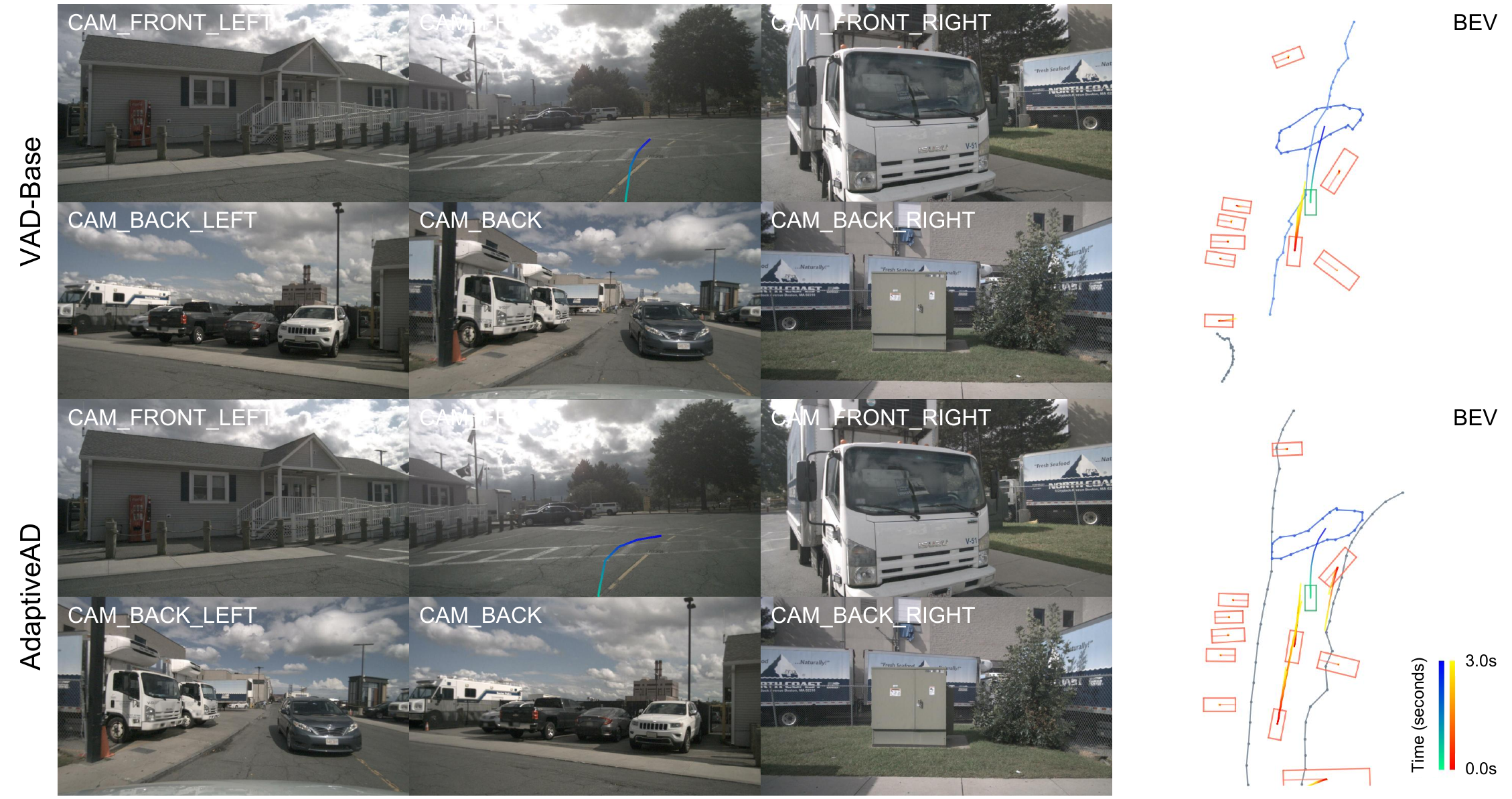}
    \caption{
        Qualitative comparison of scene generalization ability.
        In this challenging scenario, our AdaptiveAD demonstrates significantly superior perception capabilities compared to VAD, providing more reliable obstacle-avoidance paths.
    }
    \label{fig:mainresult}
\end{figure*}

\begin{figure}[t]
    \centering
    \includegraphics[width=.92\linewidth]{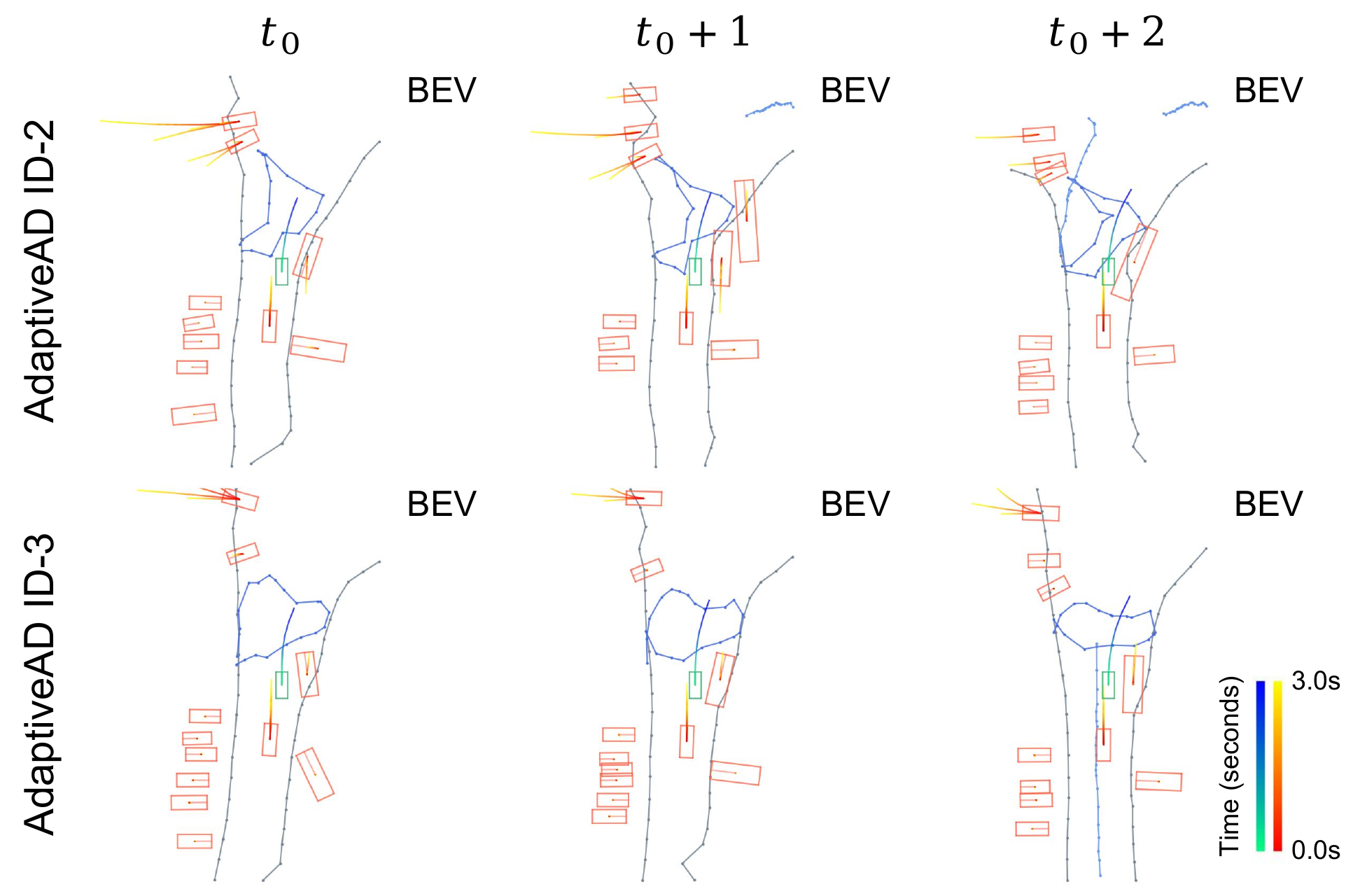}
    \caption{
        Qualitative evaluation of BEV unidirectional distillation.
    }
    \label{fig:bevdistillation}
\end{figure}

\begin{figure}[t]
    \centering
    \includegraphics[width=.92\linewidth]{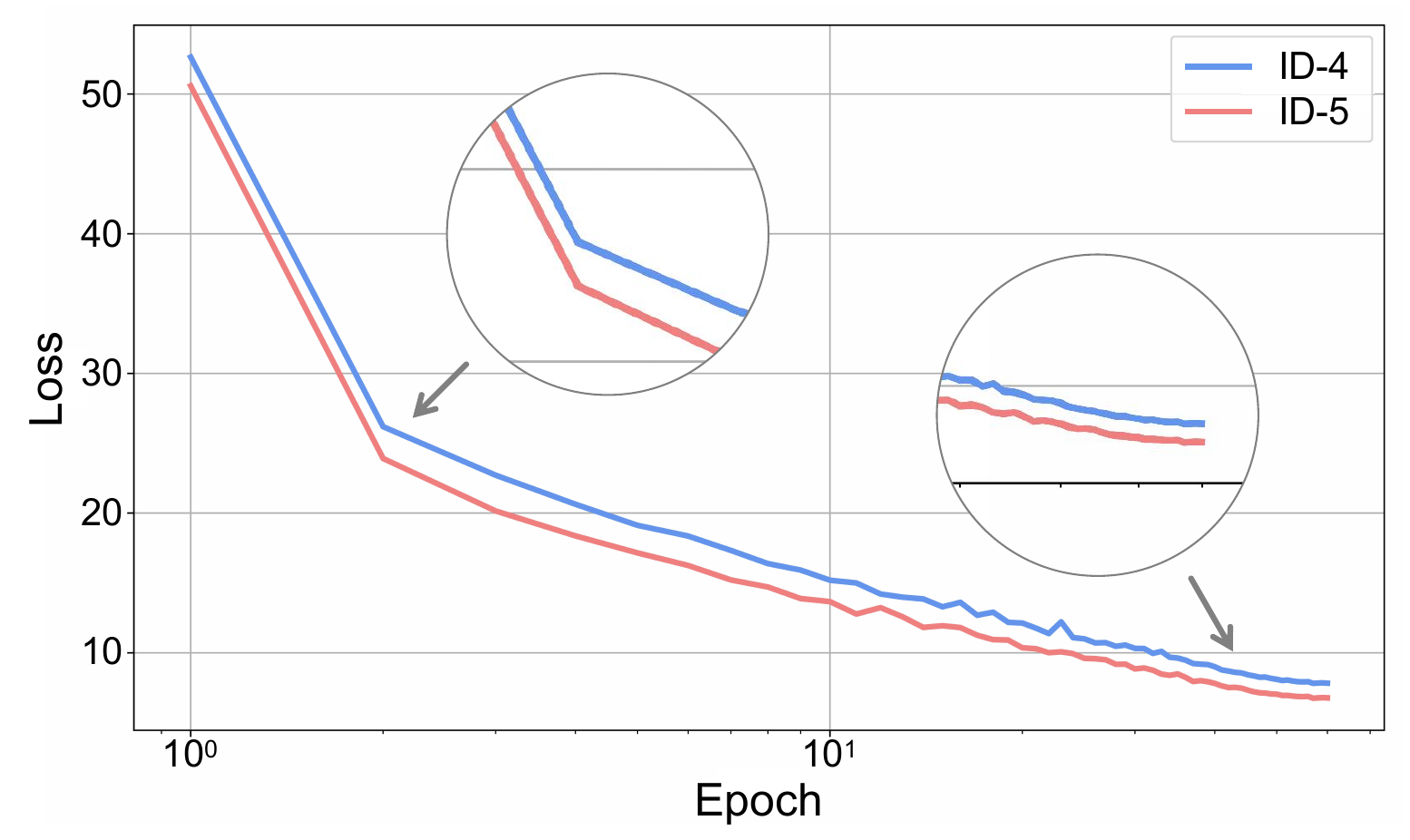}
    \caption{
        Impact of autoregressive online mapping on model convergence speed.
    }
    \label{fig:losscurves}
\end{figure}

\subsubsection{Component Contributions.}

Our baseline (ID-1), a scene-driven branch that retains ego-status enhancement, performs reasonably well. Introducing our dual-branch and fusion module but without the auxiliary regularizers (ID-2) leads to a drop in L2 accuracy, as the scene-driven branch suffers from motion blur without ego-motion compensation. The addition of BEV unidirectional distillation (ID-3) is critical; it recovers the L2 performance while dramatically reducing the collision rate by over 60$\%$. This is because distillation enhances the perceptual quality of the scene-driven BEV features ($B_{woes}$), as qualitatively verified in Figure \ref{fig:bevdistillation}. Next, incorporating scene-aware initialization (ID-4) further improves trajectory accuracy by $\sim$10$\%$, enabling the model to adaptively weight the two decision contexts based on scene complexity. Finally, adding our autoregressive online mapping task (ID-5) yields the full AdaptiveAD model, which achieves the best overall performance by improving trajectory quality without compromising safety. As shown in Figure \ref{fig:losscurves}, this auxiliary task also accelerates model convergence by mitigating the optimization conflict between the mapping and planning heads.

\subsubsection{Effectiveness of Path Attention.}

We compare our proposed path attention (P.A.) against a standard deformable attention (D.A.) mechanism \cite{zhu2020deformable} for the ego-BEV interaction module. As shown in Table \ref{tab:pathattention}, path attention achieves superior long-term planning accuracy and safety with identical computational overhead. This confirms that guiding attention sampling along the hypothesized future trajectory is a more effective strategy for this task.

\subsubsection{Generalizability of Components.}

To demonstrate the broader utility of our contributions, we integrate path attention (P.A.) and autoregressive online mapping (A.O.M.) as plug-in modules into two other SOTA models, UniAD \cite{hu2023planning} and SparseDrive \cite{sun2024sparsedrive}. As reported in Table \ref{tab:pluginplanningperformance}, both components provide consistent performance improvements, underscoring their general applicability beyond the AdaptiveAD framework.

\subsection{Qualitative Results}

Figure \ref{fig:mainresult} provides a qualitative comparison in a challenging obstacle avoidance scenario. The baseline model, likely over-relying on its prior for forward motion, fails to perceive the stopped vehicle and plans a collision course. In contrast, AdaptiveAD's robust scene understanding allows it to correctly identify the obstacle and generate a safe and comfortable avoidance maneuver. Visualizations of two additional state-of-the-art models are provided in Supplementary Material, Section C.

\section{Conclusion}
\label{sec:conclusion}

In this work, we addressed the over-reliance on ego status in end-to-end autonomous driving, diagnosing it as a potential architectural flaw rather than a mere dataset bias. We introduced AdaptiveAD, a framework that remedies this issue through a principled decoupling of scene-driven and ego-driven reasoning, followed by an adaptive fusion module. AdaptiveAD achieves state-of-the-art open-loop planning performance on the nuScenes benchmark. Crucially, our experiments demonstrate this performance improvement does not arise from overfitting to benchmark statistics, but rather from a significant suppression of the ego-state shortcut and substantially improved robustness in complex scenarios, a finding corroborated across both open- and closed-loop evaluations.

Beyond a specific implementation, our work champions a broader principle: the explicit decoupling and adaptive fusion of distinct reasoning contexts is a powerful paradigm for building more robust and generalizable driving models. This architectural philosophy opens several promising avenues for future research. For instance, the modularity of our decoupled design provides a clear path for integration with other advanced systems, such as generative world models, to further enhance causal reasoning and scene understanding. Furthermore, it offers a structured approach to enhancing training efficiency and extending the framework to handle multimodal planning. By promoting a more robust and causally sound approach to decision-making, this research contributes to the development of safer, more scalable autonomous driving systems.

\bibliography{aaai2026}

\end{document}


\maketitle

\begin{abstract}
This supplementary document provides additional details to accompany our main paper. Section A elaborates on our experimental setup, including the architecture of our baseline model, a formal definition of the optimization objectives, and a comprehensive list of implementation details necessary for reproducibility. Section B offers in-depth discussions on key methodological nuances, addressing the potential for information leakage through BEV distillation and clarifying the architectural novelty of our approach compared to prior work. Finally, Section C presents additional qualitative results, with visual comparisons that further substantiate our claims regarding improved robustness and scene understanding.
\end{abstract}


\section{A. Experimental Setup Details}

\subsection{A.1. Baseline Architecture}

Our work builds upon VAD \cite{jiang2023vad}, a state-of-the-art, planning-oriented autonomous driving framework. VAD employs a unified architecture composed of an image backbone with a feature pyramid network (FPN) \cite{lin2017feature}, a BEV encoder \cite{li2024bevformer}, a vectorized scene decoder for detection and mapping, and dedicated decoders for multi-modal motion prediction and planning. A critical characteristic of VAD, which motivates our work, is its method of fusing ego-status information. Specifically, high-dimensional features encoded from the ego vehicle's kinematic state are used to enhance BEV queries within the BEV encoder. This early and direct fusion creates a strong informational shortcut from ego status to the final planning output, which we identify as a primary source of over-reliance. VAD serves as an ideal baseline as it prominently features the architectural patterns that AdaptiveAD is designed to resolve.

\subsection{A.2. Optimization Objectives}

We adhere to the multi-task optimization framework established by VAD. The total loss function is a weighted sum of losses from different tasks, designed to jointly optimize perception, prediction, and planning. The overall optimization objective is formally defined as:
\begin{equation}
\begin{aligned}
    \mathcal{L}_{\text{total}} ={} & \lambda_{\text{det}}\mathcal{L}_{\text{det}} + \lambda_{\text{map}}\mathcal{L}_{\text{map}} && \text{(Perception)} \\
    & + \lambda_{\text{mot}}\mathcal{L}_{\text{mot}} && \text{(Prediction)} \\
    & + \lambda_{\text{plan}}\mathcal{L}_{\text{plan}} + \lambda_{\text{aux}}\mathcal{L}_{\text{aux}} && \text{(Planning \& Auxiliary)}
\end{aligned}
\end{equation}
where $\mathcal{L}_{\text{det}}$ and $\mathcal{L}_{\text{map}}$ are the losses for 3D object detection and online mapping, respectively, following a set-based Hungarian matching loss formulation \cite{carion2020end}. $\mathcal{L}_{\text{mot}}$ is the motion prediction loss, combining a classification loss for mode selection and a regression loss (L1) for trajectory waypoints. $\mathcal{L}_{\text{plan}}$ is the planning loss, which holistically evaluates planning accuracy, safety, and ride comfort. $\mathcal{L}_{\text{aux}}$ encompasses the losses for our proposed auxiliary tasks, namely BEV unidirectional distillation and autoregressive online mapping, as detailed in the main paper. All loss weights ($\lambda$) are kept consistent with the VAD framework to ensure a fair comparison.

\subsection{A.3. Implementation Details}

Our experiments are conducted using the MMDetection3D codebase. The architecture's backbone, a ResNet-50 network \cite{he2016deep} with FPN pre-trained on nuImages, is shared between the scene-driven and ego-driven branches to maintain efficiency. Input images from six cameras are resized to a resolution of 256$\times$704. The perception range covers a region of 60m in length ([-30m, 30m]) and 30m in width ([-15m, 15m]), with a BEV grid resolution of 0.15m.

For the query-based components, we set the number of agent queries ($N_{agent}$) to 300, map queries ($N_{map}$) to 100, and planning modes ($N_{mode}$) to 3. The feature dimension ($C$) for all queries and BEV features is 256. For optimization, we use the AdamW optimizer \cite{loshchilov2017decoupled} with a base learning rate of 2e-4 and a weight decay of 0.01. A cosine annealing learning rate scheduler \cite{loshchilov2016sgdr} is employed, with a linear warmup for the first 500 iterations. We follow the standard data augmentation pipeline from VAD, which includes photometric distortion and image scaling with a fixed factor. To ensure stable knowledge transfer, gradients are stopped for the teacher branch exclusively during BEV distillation.

\section{B. Discussion on Methodological Nuances}

\subsection{B.1. Analysis of Potential Information Leakage}

We acknowledge the possibility that ego-status cues could be transferred from the ``teacher'' BEV feature ($B_{wes}$) to the ``student'' ($B_{woes}$) via distillation, potentially reintroducing a subtle shortcut. However, we contend that this risk is minimal due to several integral architectural constraints. Principally, the distillation loss is spatially confined, as our agent-guided reweighting mechanism (Eq. 6 in the main paper) focuses the objective primarily on foreground regions corresponding to dynamic agents. This limits the scope of any potential leakage. Furthermore, any residual cues must traverse a highly indirect and lossy information pathway: from the distilled BEV features, through agent queries, and finally to the ego query during planner-level interactions. Additionally, our path attention mechanism inherently mitigates this risk by sampling features along the hypothesized future trajectory, which predominantly consists of navigable space rather than the dense foreground regions most affected by distillation. The above analysis reveals that, compared to the premature and direct feature fusion in VAD, any potential information transfer through distillation is negligible.

\subsection{B.2. Architectural Distinction from Prior Work}

The primary distinction between AdaptiveAD and prior works like SparseDrive \cite{sun2024sparsedrive} and BridgeAD \cite{zhang2025bridging} lies in the level of intervention used to mitigate ego-status over-reliance. The shortcut from ego status to planning in end-to-end models typically manifests in two key areas: (1) the enhancement of BEV queries with ego features in the BEV encoder, and (2) the concatenation or addition of ego features with the ego query in the planning head.

While methods like SparseDrive and BridgeAD effectively avoid the first risk due to their sparse paradigm and mitigate the second by learning ego status as an auxiliary task, this learning process may still introduce spurious correlations from dataset biases. The shortcut path in the planning head remains structurally intact, albeit weakened. In contrast, AdaptiveAD proposes a more principled, architectural-level solution. It completely removes ego-feature fusion in the planner (addressing risk 2) and focuses on resolving the more challenging BEV-level leakage (risk 1) through explicit decoupling. By physically separating the information flows for scene-driven and ego-driven reasoning, our framework suppresses the shortcut at its source. Our perturbation experiments (Tables 3 \& 4 in the main paper) validate this approach, demonstrating superior resilience when ego-status inputs are unreliable. We posit that these two strategies, learning ego status and architectural decoupling, are not mutually exclusive but are potentially complementary. Exploring their synergy is a promising direction for future work.

\begin{figure*}[t]
    \centering
    \includegraphics[width=0.9\linewidth]{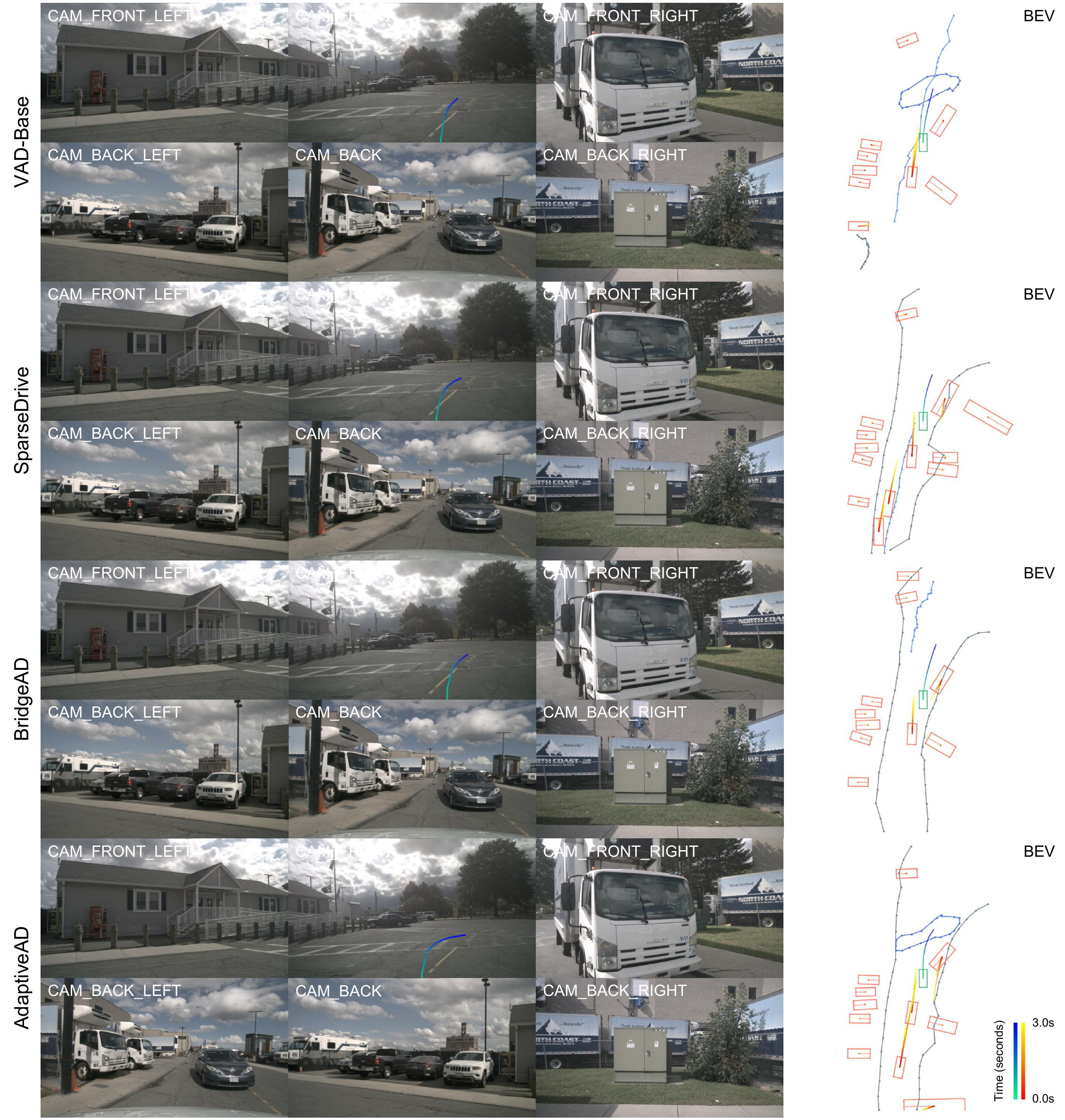} 
    \caption{
        Qualitative comparison in a challenging obstacle avoidance scenario. VAD \cite{jiang2023vad} over-relies on its motion prior and plans a collision. SparseDrive \cite{sun2024sparsedrive} and BridgeAD \cite{zhang2025bridging} perceive the obstacle but plan hesitant or suboptimal paths. Our AdaptiveAD correctly perceives the scene and generates a safe, smooth avoidance maneuver, demonstrating superior robustness.
    }
    \label{fig:supp_qualitative}
\end{figure*}

\section{C. Additional Qualitative Results}

To further illustrate the benefits of our approach, we provide an extended qualitative comparison in a challenging, long-tail scenario where a static vehicle unexpectedly blocks the ego vehicle's path. As shown in Figure \ref{fig:supp_qualitative}, the baseline VAD exhibits classic ``driving by inertia''. Over-relying on its high-velocity prior, it fails to correctly perceive the stationary obstacle and consequently plans a hazardous trajectory leading to a collision. SparseDrive and BridgeAD demonstrate improved perception over the baseline, successfully detecting the obstacle. However, their planned trajectories still exhibit hesitation or are overly conservative, potentially influenced by residual dataset priors. In stark contrast, AdaptiveAD's scene-driven branch enables robust perception of the environment, free from the bias of ego motion. This allows the model to correctly identify the obstacle and its context, leading to the generation of a smooth, decisive, and safe avoidance maneuver. This visual evidence corroborates our quantitative findings, highlighting how architecturally decoupling scene perception from ego-status priors enables more robust and reliable decision-making in safety-critical scenarios.

\bibliography{aaai2026}